\documentclass[10pt,twocolumn,letterpaper]{article}

\usepackage{cvpr}              

\usepackage{graphicx}
\usepackage{amsmath}
\usepackage{amssymb}
\usepackage{booktabs}

\usepackage{makecell}
\usepackage{enumitem}

\usepackage[pagebackref,breaklinks,colorlinks]{hyperref}

\usepackage[capitalize]{cleveref}
\crefname{section}{Sec.}{Secs.}
\Crefname{section}{Section}{Sections}
\Crefname{table}{Table}{Tables}
\crefname{table}{Tab.}{Tabs.}

\def\cvprPaperID{6} 
\def\confName{CVPR}
\def\confYear{2023}

\begin{document}

\title{Upscaling Global Hourly GPP with Temporal Fusion Transformer (TFT)}

\author{Rumi Nakagawa\thanks{These authors contributed equally.}
\and
Mary Chau$^{*}$
\and
John Calzaretta$^{*}$
\and
Trevor Keenan
\and
Puya Vahabi
\and
Alberto Todeschini
\and
Maoya Bassiouni
\and
Yanghui Kang
\and 
\\
University of California, Berkeley\\
{\tt\small \{ruminakagawa, marychau, john.calzaretta, trevorkeenan, puyavahabi,}\\ 
{\tt\small todeschini, maoya, yanghuikang\}@berkeley.edu}
}

\maketitle

\begin{abstract}

Reliable estimates of Gross Primary Productivity (GPP), crucial for evaluating climate change initiatives, are currently only available from sparsely distributed eddy covariance tower sites. This limitation hampers access to reliable GPP quantification at regional to global scales. Prior machine learning studies on upscaling \textit{in situ} GPP to global wall-to-wall maps at sub-daily time steps faced limitations such as lack of input features at higher temporal resolutions and significant missing values. This research explored a novel upscaling solution using Temporal Fusion Transformer (TFT) without relying on past GPP time series. Model development was supplemented by Random Forest Regressor (RFR) and XGBoost, followed by the hybrid model of TFT and tree algorithms. The best preforming model yielded to model performance of 0.704 NSE and 3.54 RMSE. Another contribution of the study was the breakdown analysis of encoder feature importance based on time and flux tower sites. Such analysis enhanced the interpretability of the multi-head attention layer as well as the visual understanding of temporal dynamics of influential features.

\end{abstract}

\section{Introduction}\label{sec:intro}

The ongoing increase in CO\textsubscript{2} emissions, despite a temporary slowdown during the 2020 pandemic, has mobilized government agencies, industries, and individuals to tackle climate change. As investments in climate change mitigation initiatives continue to grow, reliable estimation of carbon flux becomes essential for informed green policies, global carbon budgets, and ensuring the effectiveness and accountability of these actions. Gross Primary Productivity (GPP), which refers to the quantification of carbon uptake through vegetation, is a key factor in the carbon cycle and estimated from CO\textsubscript{2} exchanges measurements. However, the measurement of CO\textsubscript{2} exchanges is currently limited to a sparse network of flux towers, tall instrumented structures used to quantify gas exchange between the land surface and atmosphere of the Earth. With less than a thousand flux towers globally, the availability of carbon flux measurements is geographically limited. Given the extreme sparsity of sites with GPP measurements, the development of an upscaling model is imperative to estimate GPP at locations worldwide. This upscaling task is particularly challenging due to regional biases in the available flux data, mainly concentrated in North America and Europe. Improving empirical models for global inference will enhance the accessibility of quantified CO\textsubscript{2}, regardless of the availability of flux tower equipment.

Past literature that studied global GPP products has shifted its focus to higher temporal resolutions, from annual \cite{Beer2010Terrestrial}, monthly\cite{Jung2009Towards}\cite{Jung2011Global}, daily\cite{Tramontana2016Predicting}, and to the latest study on half-hourly\cite{Bodesheim2018Upscaled} level. Estimating GPP at sub-daily or hourly time scales is crucial for understanding important ecosystem-climate interactions, such as the lagged or legacy effects, especially during extreme climate events, which are projected to become more frequent under climate change \cite{bg-15-1293-2018}\cite{hess-24-5579-2020}. However, the model with half-hourly resolution encountered the limited availability of features of the same granularity. In the latest study, meteorological features were constrained to daily resolution as the highest level of detail\cite{Bodesheim2018Upscaled}. Moreover, machine learning studies in this field have not adequately captured the temporal dynamics, partially due to the computational resources required for time-aware models, such as long short-term memory (LSTM).

Temporal Fusion Transformer (TFT) possesses key characteristics that address these limitations: 1) TFT accommodates time series of different time periods. 2) TFT can forecast on previously unseen entities. 3) TFT handles diverse inputs, including heterogeneous time series, time-varying features, and static metadata. 4) TFT utilizes LSTM and self-attention mechanism to learn historical patterns in both short- and long-term. 5) TFT offers powerful interpretability. Built-in function of Pytorch allows feature importance analysis by the encoder, decoder and static features at each prediction time step. Overall, TFT potentially serves as an effective upscaling solution, and provides rich information to comprehend the behavior of the model.

This study applied TFT to upscale hourly GPP by incorporating time-aware elements. The primary objectives were to enhance the performance of the upscaling model and to analyze the temporal dynamics of influential features in the TFT model output. Additionally, Random Forest Regressor (RFR) and XGBoost (XGB) models were developed in parallel using the same dataset. Notably, three modeling approaches were employed. First, a non-upscaling capable model with past GPP values was constructed to establish the best possible performance benchmark for the TFT model. Second, a TFT model was built without the past GPP values of the test sites to serve as an upscaling solution. Finally, a two-stage model, TFT models was developed utilizing the predicted GPP values of the tree algorithms as potential alternatives of the unavailable past GPP values. It aimed to assess whether estimated past GPP values could improve to model performance than not providing any past GPP values.

\section{Literature Review}

The FLUXNET project, initiated in 1997 to develop accessible ground truth, was continued with the cooperation of scientific communities to develop networks across Europe, North America, and Asia. The first empirical model that used machine learning applied artificial neural networks to data in European forests \cite{papale2003New}. Support Vector Regression \cite{yang2007Remote}\cite{ueyama2013Upscaling}\cite{ichii2017New} and ensemble tree models\cite{xiao2008Estimation}\cite{xiao2010Continuous}\cite{Jung2009Towards}\cite{Bodesheim2018Upscaled} also have been actively studied. Tramontana et al. compared 16 machine learning algorithms, including kernel methods, neural networks, tree methods, and regression splines\cite{Tramontana2016Predicting} and reported high consistency in the model performances, although site-dependencies were observed. While recurrent neural networks, such as LSTM was applied to CO\textsubscript{2} flux prediction as a time series forecasting model \cite{Besnard2019Memory}, its application to GPP has not been explored in previous literature.

Satellite-based remote sensing features are commonly used in this field due to their geographically widespread availability and ability to indicate vegetation structural changes \cite{xiao2008Estimation}\cite{xiao2010Continuous}\cite{xiao2012Advances}. Past studies utilized remote sensing features such as  Land Surface Temperature (LST)\cite{WANValidation}\cite{wang2008Remote}, Normalized Difference Vegetation Index (NDVI), Enhanced Vegetation Index (EVI) \cite{HUETE2002195}, Leaf Area Index (LAI), fraction of absorbed photosynthetic active radiation (fPAR) \cite{MYNENI2002214}, Normalized Difference Water Index (NDWI)\cite{GAO1996257}. In addition to remote sensing features, meteorological features such as air temperature (TA), vapor pressure deficit (VPD), water availability \cite{Jung2017Compensatory} were also often included in the past studies \cite{Tramontana2016Predicting}\cite{Bodesheim2018Upscaled}. 

GPP upscaling with higher temporal resolution has gained attention in the past two decades. Previous studies by Jung et al. in 2009 and 2011 focused on monthly GPP \cite{Jung2009Towards}\cite{Jung2011Global}, while Beer et al. examined median annual GPP in 2010 \cite{Beer2010Terrestrial}, and Tramontana et al. explored daily GPP in 2016 \cite{Tramontana2016Predicting}, followed by Bodesheim et al. utilizing Random Forest Regressor (RFR) for half-hourly predictions in 2018 \cite{Bodesheim2018Upscaled}. Bodesheim et al. applied the remote sensing and meteorological features previously utilized by Tramontana et al. Although the model targets were at a half-hourly level, most features, except one radiation feature, were on daily level, resulting in common values for every 48 time steps. To overcome this limitation, Bodesheim et al. employed two approaches: 1) Building a model for each of the 48 time steps, 2) Constructing a single model with the first order derivative of target value (GPP) as a supplemental engineered feature. The latter approach resulted better model performance, with an NSE value of 0.67.

\section{Data Sources}

In order to build a global GPP upscaling model and fully utilize the ability of TFT in handling heterogeneous inputs, this study incorporated features from multiple global data sources, encompassing remote sensing and meteorological data at different temporal resolutions, including hourly, daily, 4-day, 8-day, 16-day, and monthly intervals. Additionally, climate and land classification labels for each location were included as static metadata. The target variable, GPP, was directly obtained from the FLUXNET2015 Dataset \cite{pastorello2020fluxnet2015}.

\subsection{GPP Ground-Truth: FLUXNET}

FLUXNET is a valuable source of original and processed data collected from flux towers, with the latest data product, FLUXNET2015 Dataset, hosted by Lawrence Berkeley National Laboratory. Carbon fluxes are measured through flux towers using the eddy covariance method, which quantifies gas exchange between the biosphere and the atmosphere. FLUXNET offers four types of GPP based on the timing of respiration measurements, and the calculation methods for thresholds. GPP-NT-VUT-REF was defined as the target GPP variable in this study. The duration of available GPP data varies drastically across sites, ranging from over 20 years to just a few weeks (\Cref{fig:dataset-distribution-by-sites}).

\subsection{Global Features: Remote-Sensing Data}


MODIS (Moderate Resolution Imaging Spectroradiometer) is the key instrument carried by the Terra (EOS AM-1) and Aqua (EOS PM-1) satellites. These satellites capture images of the surface of the Earth every day or every two days, providing data with a spatial resolution of 500 m to 1 km \cite{JUSTICE20023}. The study used the following MODIS datasets with different resolutions and features: 

\begin{enumerate}[nosep]
  \item MCD43C4 (Daily): NIRv (Near-Infrared Reflectance of Vegetation\footnote{canopy structure that are measured in remote sensing}), NDVI, EVI, NDWI, percentage of snow, PET (Evapotranspiration), Surface reflectance b1 to b7 \cite{MCD43C4} 
  \item MCD15A3H (4-day): LAI (Leaf Area Index), fPAR (Fraction of Photosynthetically Active Radiation) \cite{MCD15A3H} 
  \item MCD12Q1 (Annual): MODIS-IGBP (International Geosphere–Biosphere Programme) and PFT (Plant Functional Type) \cite{MCD12Q1}
  \item MYD11A1 (Daily): Daytime LST (Land Surface Temperature), Nighttime LST \cite{MYD11A1}
\end{enumerate}

Two other global remote sensing data were applied in this study: 1) Continuous Solar-Induced Fluorescence (CSIF) with 4-day resolution derived from Orbiting Carbon Observatory-2 (OCO-2) SIF observations and MODIS surface reflectance. Inclusion of CSIF potentially allows models to capture more details on the diurnal and seasonal patterns of photosynthesize activities. 2) Photosynthetically active radiation (PAR), diffuse PAR, and shortwave downwelling radiation (RSDN) derived from Breathing Earth System Simulator (BESS) models. These BESS\_Rad data helps account for variations in light availability and quality that can affect photosynthesis in different vegetation types and canopy structures, thereby potentially contributing to estimates of GPP.
 
\subsection{Global Features: Meteorological Data}
Two notable meteorological and climate data sources in the study are ERA5-Land and Köppen-Geiger classification. ERA5-Land is a dataset from the European Centre for Medium-Range Weather Forecasts (ECMWF), offering hourly climate and global environmental data on land surface \cite{ERA5MuñozSabater}. Air temperature, VPD, precipitation, skin temperature, soil moisture, potential evapotranspiration, shortwave and longwave radiation from this dataset are used as features in this study. The Köppen classification represents the major climate groups base on annual temperature and precipitation patterns while Köppen-sub is a subcategory with more details on seasonality and precipitation characteristics.

\section{Models and Method}
This study focused on applying a temporal fusion transformer (TFT), supplemented by RFR and XGBoost regression to develop GPP upscaling models.

Temporal Fusion Transformer (TFT), introduced by Google Cloud AI in 2019 \cite{lim2020temporal}, is an attention-based time series forecasting model that has outperformed existing models in various time series forecasting tasks \cite{lim2020temporal}\cite{WUInterpretable}. The unique components in TFT architecture make it a powerful and interpretable model for time series forecasting. It applies a static covariate encoder to integrate static metadata into the temporal fusion decoder, enabling temporal forecasts conditioned on such information. Its gating components allow the model to skip unnecessary parts of the network, improving model efficiency. Variable selection is applied to both time-variant and time-invariant input features, selecting relevant features and removing noisy inputs to enhance model performance. Moreover, the model learns short- and long-term patterns through sequence-to-sequence (LSTM) and attention-based layers, identifying temporal dynamics such as seasonality and significant events.

Given $I$ unique entities in a time series data, each entity $i$ has associated static covariates $\textbf{s}_{i}\in \mathbb{R}^{m_{s}}$, inputs features $\boldsymbol\chi_{i,t} \in \mathbb{R}^{m_{x}}$ and scalar targets $y_{i,t} \in \mathbb{R}$ at each time step $t$. The input features are subdivided into two categories: 1) observed inputs $\textbf{\textit{z}}_{i,t} \in \mathbb{R}^{m_{x}}$ that are known prior time $t$ but unknown afterward 2) predetermined inputs $\textbf{\textit{x}}_{i,t} \in \mathbb{R}^{m_{x}}$ which remains known after time $t$ (e.g. the hour of the day) . Each quantile forecast can be represented as:
\begin{equation}\label{tft-equation}
	\hat{y}_{i}(q,t,\tau) = f_{q}(\tau, y_{i,t-k:t},\textbf{\textit{z}}_{i,t-k:t},\textbf{\textit{x}}_{i,t-k:t+\tau},\textbf{s}_{i})
\end{equation}
where $\hat{y_{i}}(q, t, \tau) $ is the predicted $q^{th}$ sample quantile of the $\tau$-step-ahead forecast at time $t$ and $f_{q}(.)$ is the TFT prediction model. The model incorporates the past target values and observed values within a finite look-back window of length $k$ (i.e., $y_{i,t-k:t}= {y_{i,t-k},...,y_{i,t}}$), followed by known input features across the entire time range (i.e., $\textbf{\textit{x}}_{i,t-k:t+\tau}= {\textbf{\textit{x}}_{i,t-k},...,\textbf{\textit{x}}_{i,t},...,\textbf{\textit{x}}_{i,t+\tau}}$), to produce forecasts for $\tau_{max}$ time steps ahead.

In the context of reconstructing past global GPP, each location (e.g. flux tower) corresponded to an entity $i$. All input features except GPP measurements were available in prediction time $t$ as well as $\tau$ steps ahead. Remote sensing and meteorological data, varying by time, were considered as time-varying known features. Likewise, time-related information such as month, day, and hour of the record fell in the same input type as well. The metadata of each location, such as IGBP and Köppen labels, were treated as static covariates. Since the study focused on single-point estimate of the past GPP at unseen sites, the decoder length ($\tau$) of the TFT models were consistently set to 1. The encoder length ($k$) was a hyperparameter to be optimized to balance model complexity and performance.

\subsection{Evaluation metrics}
Three evaluation metrics were selected to assess the performance of the GPP upscaling models and to compare with the past literature: 1) Root Mean Squared Error (RMSE), 2) Mean Absolute Error (MAE), and 3) Nash-Sutcliffe Efficiency (NSE). NSE is a normalized statistic that determines the relative magnitude of the residual variance in comparison with the measured data variance \cite{nash1970River}. Although originally developed for hydrology, the NSE is commonly employed in upscaling tasks and is mathematically equivalent to the coefficient of determination ($R^{2}$) \cite{Jung2011Global}\cite{Tramontana2016Predicting}\cite{Bodesheim2018Upscaled}. 

\section{Data Preprocessing}

\begin{figure}
    \centering
    \includegraphics[width=\linewidth]{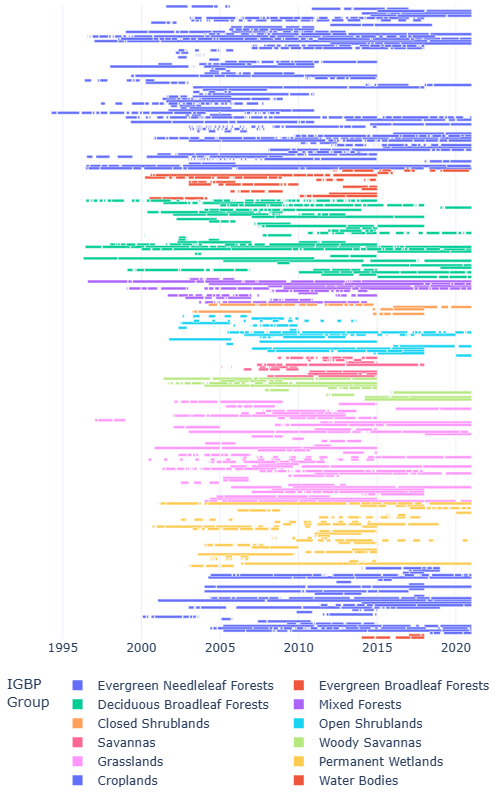}
    \caption{Timelines of available observations of all flux towers (274 total) with GPP measurement, organized by IGBP groups. Each horizontal line corresponds to a single flux tower site, and colored area depicts the available period of data in each site.}
    \label{fig:dataset-distribution-by-sites}
\end{figure}

To accommodate time and computational constraints, the modeling period for the data was set from 2010 to 2015, during which a greater number of observations and relevant sites were accessible. As depicted in \Cref{fig:dataset-distribution-by-sites}, the number of active flux towers significantly decreased after 2015. In order to mitigate the impact of gap-filling and to preserve the year-long seasonality in the dataset, the flux towers with less than a year of available data and/or over 20\% of missing records were excluded. The resulting dataset applied to the modeling comprised 129 flux tower sites.

\subsection{Imputation and Gap-Filling}

After the initial filtering of the flux tower sites, the dataset still had a considerable amount of missing values and gaps (i.e. time steps with no records). The gaps in record sequences present a challenge for time series models, such as TFT, that require continuous sequences of input data.

K-nearest neighbors (KNN) imputation was employed to address the challenge of large amount of missing data. 
This approach considers temporal dependencies, fills large blocks of missing values, and maintains time series continuity by identifying similar neighbors based on Euclidean distance, thereby making it well-suited to address the problem at hand. Two KNN imputation methods were used in the study: one for filling missing values within time step records and another for filling gaps in record sequences. Both methods utilized 5 neighbors, Euclidean distance, and a uniform average of neighbor values, determined through testing and validation with the RFR model. Gap-filling flags were included as observed input features in all TFT experiments to indicate whether a record was gap-filled or not.

\subsection{Stratified Train/Test Split}
Since the number of flux towers is limited, it is vital to maintain diversity in training, validation, and test sets to ensure generalization and objective evaluation of the model. In order to realize the diverse distribution of each dataset, stratified data split was employed by generic IGBP groups. It combined some categories, such as Open and Closed Shrublands into Shrublands, and Woody Savannas and Savannas into Savannas. This stratified approach resulted in 78 sites for training, 26 sites for validation, and 25 sites for testing.

\section{Experimental design}

\begin{table}[ht]
\centering
\begin{tabular}{>{\centering\arraybackslash}m{0.05\linewidth} m{0.25\linewidth} m{0.225\linewidth} >{\centering\arraybackslash}m{0.2\linewidth}}
\toprule
Step & Experiment & Algorithms & Upscaling Capable \\ \hline
\midrule
1 & Baseline & RFR & Yes \\ \hline
2 & \vspace{0.1\topsep}Feature-Engineered Trees & \vspace{-0.5\topsep}\makecell[l]{RFR\\XGBoost} & Yes \\ \hline
\vspace{0.2\topsep}3 & \vspace{0.2\topsep}GPP-TFT & \vspace{0.2\topsep}TFT & \vspace{0.2\topsep}No \\ \hline
\vspace{0.2\topsep}4 & \vspace{0.2\topsep}No-GPP-TFT & \vspace{0.2\topsep}TFT & \vspace{0.2\topsep}Yes \\ \hline
\vspace{0.75\topsep}5 & \vspace{0.75\topsep}Tree-FT & \vspace{0.2\topsep} \makecell[l]{Hybrid\\(Tree + TFT)} & \vspace{0.75\topsep}Yes \\ \hline
\bottomrule
\end{tabular}
\caption{Summary of 5-step experimental design.}
\label{table:experiment_summary}
\end{table}

The experimental design consisted of the 5 steps shown in \Cref{table:experiment_summary}. The first step involved building a  baseline was built using RFR and full features. Cross-validation (CV) was conducted with RFR to assess the potential volatility in model performance on different combinations of sites in training, validation, and testing sets. Due to the substantial resource demands associated with TFT model development accompanied by CV, the impact of relying a specific split group for model validation was investigated using the baseline RFR model. Subseuqently, feature engineering was applied to the tree models (RFR, XGBoost) to study whether certain features contributed significantly to the performance of non-time-aware models, followed by the dimensionality reduction. The output of tuned tree models was utilized in the hybrid model for subsequent experiments.

The initial experiment on the GPP-TFT model followed the default TFT usage, assuming the ideal scenario where past GPP values were available at prediction time. Though the ultimate goal of the study was upscaling, a scenario where historical GPP measurements would not be available at locations without flux towers, the best-possible benchmark model was developed to understand the upper limit of the TFT model performance.  The Tree-FT models substituted historical GPP input with estimated past GPP values from the tree models, while the No-GPP-TFT models were developed without any past GPP provided as input features. Additionally, the study also explored temporal dynamics of model behavior by analyzing interpretable outputs of the TFT model.

\section{Results}

\begin{figure}
    \centering
    \includegraphics[width=\linewidth]{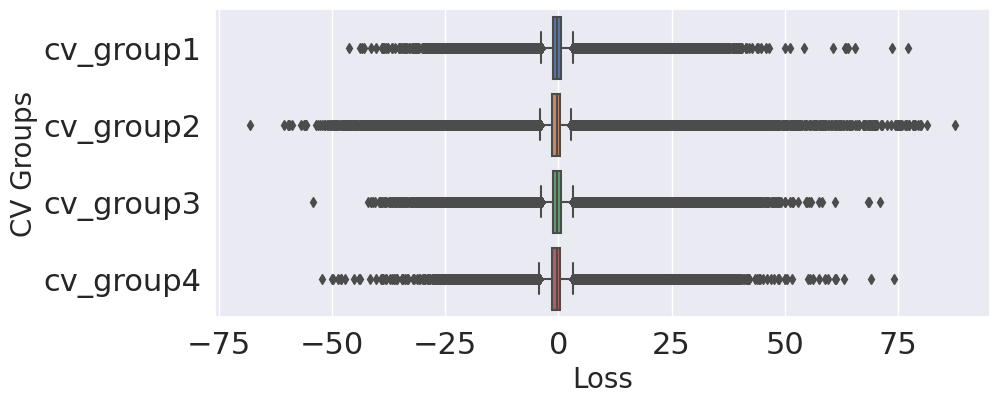}
    \caption{Distribution of loss of each CV group on RFR-BASELINE.}
    \label{fig:cv_distribution_of_loss}
\end{figure}

 Data prepocessing with stratified train/test split divided the flux tower sites into five groups. Reserving one group for testing, 4-fold CV were performed on the tree models. RFR model performance (\Cref{table:cv-group-distribution}) and  distributions of loss (\Cref{fig:cv_distribution_of_loss}) across the CV groups showed negligible differences among the four groups, suggesting that using one specific CV group as validation set throughout the study may be sufficient for objective model evaluation. Based on these findings, the TFT models was developed with consistently assigning fourth CV group as the validation set.

\begin{table}
\centering 
\begin{tabular}{lccc}
\toprule
CV Group & RMSE & MAE & NSE   \\ \hline
\midrule
CV Group1 & 3.657 & 1.979 & 0.681 \\ \hline
CV Group2 & 3.644 & 1.963 & 0.683 \\ \hline
CV Group3 & 3.650 & 1.975 & 0.682 \\ \hline
CV Group4 & 3.675 & 2.011 & 0.678 \\ \hline
\bottomrule
\end{tabular}
\caption{RFR-BASELINE model performance of each CV group.}
\label{table:cv-group-distribution}
\end{table}

The results of the best models after hyperparameter tuning are presented in \Cref{table:results_comparison}. Model 1 was the baseline using RFR, and Model 2 and 3 implemented dimensionality reduction, with each algorithm having the optimal number of features. Model 4, GPP-TFT, included past GPP values as a predictor feature. Models 5 and 6 were Tree-FT models, which were upscaling models that utilized the predicted GPP values from RFR and XGBoost as estimated past GPP values. Model 7, the No-GPP-TFT model, represented the upscaling approach without any past GPP value.

In summary, the best-performing upscaling model was the RFR model with the top 9 features (Model 2) with 3.423 RMSE and 0.704 NSE. The GPP-TFT model, though not capable of upscaling, achieved 2.132 RMSE And 0.886 NSE, implying the potential performance ceiling for global hourly GPP upscaling models. As a result of hyperparameter tuning and feature engineering, Tree-FT and No-GPP-TFT models, both capable of upscaling, also demonstrated improvements when compared to the past literature.


\begin{table*} [ht]
\centering
\begin{tabular}{p{0.28\linewidth} p{0.13\linewidth} >{\centering\arraybackslash}p{0.08\linewidth} >{\centering\arraybackslash}p{0.05\linewidth} >{\centering\arraybackslash}p{0.05\linewidth} 
>{\centering\arraybackslash}p{0.06\linewidth} >{\centering\arraybackslash}p{0.05\linewidth} cccc}
\toprule
Upscaling Model & Features &  Hidden Size & Encoder Length & Decoder Length & RMSE & MAE & NSE \\
\midrule
\textbf{Past Literature} &  &  &  &  & \\
0.  Bodesheim 2018(RFR) & - & -  & - & - & 3.940 & - & 0.670 \\
\textbf{Baseline} &  &  &  & \\
1.  RFR-BASELINE & Original & -  & - & - & 3.675 & 2.011 & 0.678 \\
\textbf{Feature-Engineered RFR/XGB} &  &  &  &  & \\
2.  RFR-TOP9     & Top 9 features$^{1}$ & - & - & - & 3.523 & 1.841 & 0.704 \\
3.  XGB-TOP3     & Top 3 features$^{2}$ & - & - & - & 3.610 & 1.870 & 0.677 \\
\textbf{GPP-TFT(non-upscaling)} &  & & &\\
4.  GPP-TFT-14EN-ORG$^{3}$     & Original &  136 & 24*14 & 1 & 2.132 & 1.016 & 0.886 \\
\textbf{Tree-FT} &  &  & & &\\
5.  RFR-TFT-14EN-SLIM   & Slim Features$^{4}$  & 16 & 24*14 & 1 & 3.630 & 1.900 & 0.671 \\
6.  XGB-TFT-14EN-SLIM   & Slim Features & 16 & 24*14 & 1 & 3.807 & 2.002 & 0.638 \\

\textbf{No-GPP-TFT} &  &  &  & &\\
7. No-GPP-TFT-7EN-SLIM & Slim Features & 16 & 24*7 & 1 & 3.594 & 1.904 & 0.677 \\
\bottomrule
\end{tabular}
\caption{ Upscaling model results of the test set.\\
\text {$^{1}$}TOP9: SW-IN-ERA (Shortwave Radiation), NDVI, NIRv, hour, LAI, TA-ERA, VPD-ERA (Vapor Pressure Deficit), EVI, CSIF-SIFdaily \\ {$^{2}$}TOP3: NDVI, NIRv, SW-IN-ERA \\{$^{3}$}Limited to one year of training data due to resource constraints. \\ {$^{4}$}Slim Features: TA-ERA, SW-IN-ERA, LW-IN-ERA (Longwave Radiation), VPD-ERA, P-ERA, PA-ERA, NDVI, b2, b4, b6, b7, BESS-PARdiff, CSIF-SIFdaily, ESACCI-smi, Percent-Snow, LAI, LST-Day, LST-Night}
\label{table:results_comparison}
\end{table*}

\section{TFT Model Interpretability}

TFT offers interpretability beyond the feature importance provided in tree models. It allows for explorations of feature significance over time and provides attention information that identifies the most influential encoder time step at the site level. By leveraging the interpretable model output of the TFT model, model behavior were analyzed by IGBP groups, and a novel visual-analytical approach was applied to investigate the temporal dynamics of influential features in each prediction.

\subsection{Analysis by IGBP Groups}

\begin{table}[ht]
\centering 
\begin{tabular}{m{0.5\linewidth} m{0.08\linewidth} m{0.08\linewidth} m{0.08\linewidth}}
\toprule
IGBP & RMSE & MAE & NSE \\ \hline
\midrule
Deciduous Broadleaf Forests (DBF)  & 3.315 & 1.742 & 0.859 \\ \hline
Grasslands (GRA) & 2.939 & 1.525 & 0.733 \\ \hline
Mixed Forests (MF) & 3.441 & 2.154 & 0.650 \\ \hline
Evergreen Needleleaf Forests (ENF) & 4.025 & 2.211 & 0.634 \\ \hline
\makecell[l]{Savanas\\(Woody savannas (WSA))} & \vspace{\topsep}2.933 & \vspace{\topsep}1.632 & \vspace{\topsep}0.595 \\ \hline
Wetlands (WET) & 3.900 & 2.126 & 0.571 \\ \hline
Cropland (CRO) & 5.246 & 2.941 & 0.536 \\ \hline
\makecell[l]{Shrublands\\(Open Shrublands (OSH))} & \vspace{\topsep}1.252 & \vspace{\topsep}0.698 & \vspace{\topsep}0.193 \\ \hline
Evergreen Broadleaf Forests (EBF)  & 4.391 & 2.607 & 0.050 \\ \hline
\bottomrule
\end{tabular}
\caption{ No-GPP-TFT model evaluation metrics breakdown by IGBP groups (ordered by NSE values).}
\label{table:nogpp-igbp-eval}
\end{table}

\begin{figure}[hb]
    \centering
    \includegraphics[width=\linewidth]{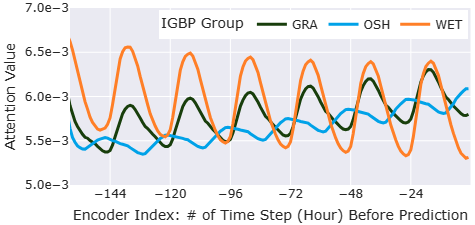}
    \caption{Attention of selected IGBP groups: Grasslands (GRA), Open Shrublands (OSH), and Wetlands (WET), obtained from the No-GPP-TFT model.}
    \label{fig:attn-by-igbp}
\end{figure}

Achieving a highly generalized and accurate GPP inference model across diverse global regions presents challenges. Evaluation metric breakdown by IGBP groups on the No-GPP-TFT model (\Cref{table:nogpp-igbp-eval}) shows variations in model performance. Deciduous Broadleaf Forests (DBF) and Grasslands (GRA) show better performance than the original model(Model 7 in \Cref{table:results_comparison}), while some IGBP groups, such as Evergreen Broadleaf Forests (EBF) and Shrublands (OSH), exhibit underperformance. The attention plot for selected IGBP groups (GRA, OSH, and WET) in \Cref{fig:attn-by-igbp} indicates potentially distinct patterns among IGBP groups. OSH displays a relatively flat attention curve that peaks around 24 hours prior, while GRA and WET show distinct daily variations peaking around 18 hours ahead. These findings suggest potential distinctive trends and differential processing of historical information based on land cover types during inference. Additionally, \Cref{fig:attn-by-igbp} highlights that attention moves in different directions over time; WET has upward trend, whereas GRA and OSH show downward trends of attention. These observations implies the importance of considering varying timescales for various regions and selecting an appropriate encoder length for the better performance of the inference.

\subsection{Temporal Analysis of Feature Importance}

\begin{figure*}[htp]
    \centering
    \includegraphics[width=\linewidth]{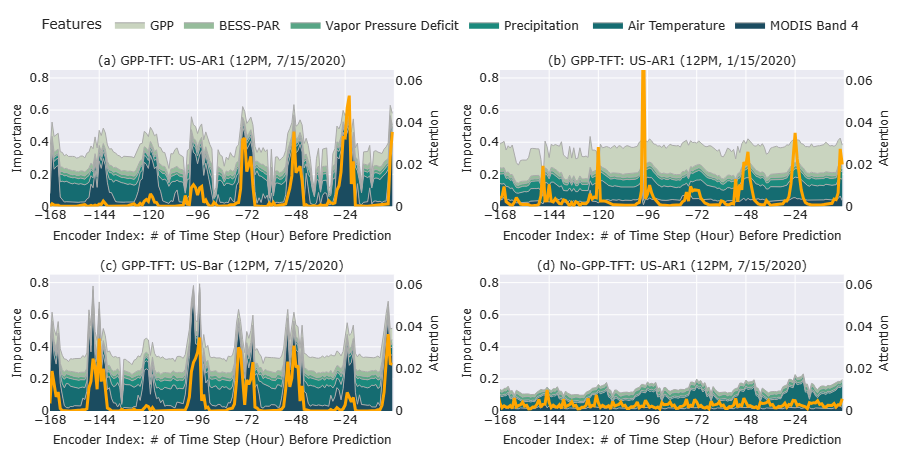}
    \caption{Temporal analysis of encoder feature importance value by encoder index on selected features. (a) GPP-TFT: Prediction at US-AR1 on July 15, 2020, at 12 PM (b) GPP-TFT: Prediction at US-AR1 on January 15, 2020, at 12 PM (c) GPP-TFT: Prediction at US-Bar on July 15, 2020, at 12 PM (d) No-GPP-TFT: Prediction at July 15, 2020, at 12 PM}
    \label{fig:tft-analysis}
\end{figure*}

There were two reasons for identifying encoder variable importance as the scope of the analysis: 1) The raw output of encoder/decoder variable importance allowed for further breakdown and examination. 2) Since the study defined decoder length as one, the resulted decoder attention was zero for all the experiments. Going beyond the default PyTorch API usage, model dynamics and behavior were analyzed by visualizing temporal allocation of feature importance. The visual provided a detailed breakdown of snapshots at each prediction time and site, which had not been extensively investigated before. By recognizing the pivotal role of encoder variables in shaping encoder attention patterns, the study meticulously plotted the evolving feature importance across the relative time indices of the encoder length. 

\subsubsection{Visual Analysis Walk-Through}
\Cref{fig:tft-analysis}a provides a snapshot of the encoder feature importance at site US-AR1\footnote{ARM USDA UNL OSU Woodward Switchgrass 1, Oklahoma, USA} of the GPP-TFT model (Model 4 in \Cref{table:results_comparison}) on July 15, 2020, with a prediction time of 12 PM. The x-axis represents hours leading up to the prediction time $t$, with negative values indicating hours before $t$ (Ex. -24 represents 24 hours prior to $t$). The left y-axis shows the feature importance value, while the right y-axis represents the attention at each encoder index. The attention of the snapshot is depicted as a bold yellow line, and the feature importance of selected features is displayed in a stacked plot format. While the \Cref{fig:tft-analysis} only covers six selected sample features, the importance are added up to one as a total of all the features in each encoder index. In more details, the top 15 influential features are listed in \Cref{table:top-features-importance}.

In this snapshot, MODIS spectral band 4 (b4; darkest green) stood out as the most influential feature. It exhibits spikes every 24 hours before the prediction time $t$ (noon in July in the northern hemisphere), while features such as air temperature (TA-ERA) peaked during the evening. The largest spikes in overall attention coincided with the spikes in b4 implies its influence in predicting GPP at $t$. The feature importance of b4 even surpasses the importance of past GPP measurements during the daytime. These patterns of b4 that GPP-TFT model captured potentially aligns properties of vegetation in biology; The relation of b4 with photosynthetic activities of vegetation\cite{YinBroadband2022}. 

\begin{table}[ht]
\centering
\setlength{\tabcolsep}{4pt}
\begin{tabular}{p{0.42\linewidth} r r r r}
\toprule
TFT Model &  \multicolumn{3}{c}{GPP-TFT} & \multicolumn{1}{c}{No-GPP} \\
Site &  \multicolumn{2}{c}{US-AR1} & \multicolumn{1}{c}{US-Bar} & \multicolumn{1}{c}{US-AR1} \\
Month & July & Jan & July & July\\
Subplots of Figure 4 & (a) & (b) & (c) & (d)  \\ \hline
\midrule
1. b4 & 11.7 & 3.9 & 14.1 & 2.3 \\ \hline
2. Hourly Gap Flag & 11.4 & 9.5 & 12.3 & 9.4\\ \hline
3. Relative Time Index & 10.4 & 12.6 & 5.4 & 1.0\\ \hline
4. GPP & 8.5 & 15.0 & 7.9 & - \\ \hline
5. Air Temperature & 8.2 & 8.3 & 8.5 & 7.3 \\ \hline
6. Global Time Index & 3.2 & 1.0 & 2.7 & 2.9 \\ \hline
7. Precipitation & 3.1 & 4.7 & 3.6 & 1.0 \\ \hline
8. LAI  & 3.0 & 1.8 & 2.1 & 2.1 \\ \hline
9. NDWI  & 2.8 & 2.9 & 2.8 & 2.3 \\ \hline
10. fPAR  & 2.6 & 3.6 & 2.0 & 2.9 \\ \hline
11. BESS-PAR  & 2.4 & 2.0 & 2.5 & 1.5 \\ \hline
12. Shortwave Radiation & 2.3 & 1.3 & 2.5 & 7.7 \\ \hline
13. PA  & 2.3 & 1.9 & 2.9 & 2.3 \\ \hline
14. PET  & 2.2 & 0.9 & 1.9 & 0.5 \\ \hline
15. b1  & 2.0 & 3.0 & 1.9 & 1.8 \\ \hline
\bottomrule
\end{tabular}
\caption{Top 15 average encoder feature importance (\%)  of snapshots in \Cref{fig:tft-analysis}. Features are ordered by the rank of \Cref{fig:tft-analysis}a, GPP-TFT model prediction at US-AR1 on July 15, 2020, at 12 PM)}
\label{table:top-features-importance}
\end{table}


\subsubsection{Comparative Analysis of Seasonality, IGBP Groups and Models} 

Comparative analysis of temporal dynamic of encoder feature importance were examined from three perspectives: 1) Seasonality 2) IGBP Groups 3) GPP-TFT versus No-GPP-TFT (Model 4 and 7 in \Cref{table:results_comparison}). When comparing the visualization of encoder feature importance between summer and winter (Figure \ref{fig:tft-analysis}a and Figure \ref{fig:tft-analysis}b), significant differences are observed in the shape and order of feature. In winter, b4 shows less impact than summer, while past GPP and air temperature exhibit increased influence. 


The feature importance was also compared at two locations representing different IGBP groups: US-AR1 (GRA, Grasslands) and US-Bar\footnote{Bartlett Experimental Forest, New Hampshire, USA} (DBF, Deciduous Broadleaf Forests) (\Cref{fig:tft-analysis}a and \Cref{fig:tft-analysis}c). The feature importance of b4 of the two locations show a shared diurnal pattern on the same prediction date and time. Similar patterns are also observed at FI-Hyy\footnote{Hyytiala, Finland} (ENF, Evergreen Needleleaf Forests) on the same prediction timestamp, and AU-DaP\footnote{Daly River Savanna, Australia} (GRA, Grasslands) on January 15, 2020 (i.e. summer at Australia). Despite the distinct climate and vegetation among different IGBP groups, similar temporal dynamics is observed. On the other hand, the model exhibits distinct trend in attention by IGBP (Yellow line in \Cref{fig:tft-analysis}a and \Cref{fig:tft-analysis}c). In US-AR1, the attention reaches the peak 24 hours before the prediction time, follow by the diminishing daily trend. It implies that day before the prediction time is the most contributing information of the prediction. On the flip side, attention is relatively evenly distributed in US-Bar, while disregarding information in one and five days prior. These differences indicate that the TFT model has the potential to capture both shared and location-specific temporal dynamics in feature importance for each prediction.

Furthermore, a comparison was made between the temporal encoder feature importance of the GPP-TFT model (\Cref{fig:tft-analysis}a) and the No-GPP-TFT model (\Cref{fig:tft-analysis}d). The key distinction between the two models lies in whether the model incorporates past GPP values as input features or not, resulting in significant differences in both model performance (\Cref{table:results_comparison}) and the ranking of influential features (\Cref{table:top-features-importance}). The selected six features in the No-GPP-TFT model (\Cref{fig:tft-analysis}d) are limited to approximately 20\% of the total importance in the snapshot. Shortwave radiation accounts for 7.7\% in the No-GPP-TFT model, whereas b4, the most impactful feature in summer snapshot of GPP-TFT model(\Cref{fig:tft-analysis}a (\Cref{fig:tft-analysis}c), is limited to 2.3\% of the importance. Similarly, relative time index, the potential clue for capturing temporal patterns, represents 1.0\% of importance in the No-GPP-TFT model, while it surpasses 10\% in the GPP-TFT model. These differences show the varying decision-making processes and model priorities based on the model setup.

\section{Conclusion}

This study addressed the crucial need for reliable estimation of GPP worldwide. Time-aware models leveraging the Temporal Fusion Transformer (TFT) model were developed to estimate hourly GPP values. Several modeling approaches were explored, including establishing a benchmark performance of TFT models with past GPP values as predictors, building TFT models without past GPP as input for upscaling, and developing two-stage models that integrated predictions from the tree models into TFT models. Comparing to the benchmark performance established in the previous work, improved performance was observed in several models. The best-performing model, feature-engineered RFR, achieved NSE of 0.704 and RMSE of 3.54.

The extensive analysis of model performance by IGBP groups showed widely distributed performance, implying distinct model dynamics by vegetation types. Instead of relying on a single common model for all geographical locations, training distinct models for each unique IGBP group might enhance the limited performance of the TFT model.  

The successful implementation of TFT in an unconventional application, solving time series forecasting problems without past target values, lays the groundwork for future upscaling solutions using TFT. Furthermore, the novel visual-analytical approach derived from interpretable outputs from TFT provides valuable insights into the temporal dynamics of feature importance, extending its potential beyond carbon flux research. These findings contribute to the advancement of GPP estimation, and hold promise for potential applications in other industries.

{\small
\bibliographystyle{ieee_fullname}
\bibliography{egbib}
}

\end{document}